\documentclass[letterpaper, 10 pt, conference]{ieeeconf}  %

\IEEEoverridecommandlockouts                              %

\usepackage[english]{babel} %
\usepackage{graphics} %
\usepackage{amsmath} %
\usepackage{amssymb}  %
\usepackage{url} %
\usepackage{tabularx} %
\usepackage{tikz} %
\usepackage{pgfplots} %
\usepackage{booktabs} %
\usepackage{caption} %
\usepackage[font=footnotesize]{subcaption} %
\usepackage{multirow} %
\usepackage{dsfont} %
\usepackage{relsize} %
\usepackage{siunitx}
\DeclareSIUnit\pixel{px}
\usepackage[capitalize]{cleveref}
\crefname{table}{Tab.}{Tabs.}
\crefname{section}{Sec.}{Sections}

\DeclareMathOperator*{\argmin}{arg\,min}
\DeclareMathOperator{\indicator}{\mathds{1}}
\def\transpose{^\top} %
\newcommand{\norm}[1]{\left\lVert#1\right\rVert}
\newcommand{\mat}[1]{\mathbf{#1}}
\newcommand{\state}{\mat{x}}
\newcommand{\states}{\mathcal{X}}
\newcommand{\statePred}{\mat{x}_{\text{pred}}}
\newcommand{\statesPred}{\mathcal{X}_{\text{pred}}}
\newcommand{\xPred}{x_{\text{pred}}}
\newcommand{\yPred}{y_{\text{pred}}}
\newcommand{\thetaPred}{\theta_{\text{pred}}}
\newcommand{\fCE}{f_{\text{CE}}}
\newcommand{\fMSE}{f_{\text{MSE}}}
\newcommand{\LCE}{L_{\text{CE}}}
\newcommand{\LMSE}{L_{\text{MSE}}}
\newcommand{\stateRef}{\mat{x}_{\text{ref}}}
\newcommand{\statesRef}{\mathcal{X}_{\text{ref}}}
\newcommand{\cells}{\mathcal{C}}
\newcommand{\cellPred}{c_{\text{pred}}}
\newcommand{\weights}{\mathcal{W}}
\newcommand{\gammaCE}{\gamma_{\text{CE}}}
\newcommand{\gammaMSE}{\gamma_{\text{MSE}}}

\newcommand{\probPath}{p_{\text{path}}}
\newcommand\copyrighttext{
	\footnotesize This work is an extended version of \cite{banzhaf2019learning} and therefore partially copyrighted by IEEE: \textcopyright~2019 IEEE. Personal use of this material is permitted. Permission from IEEE must be obtained for all other uses, in any current or future media, including reprinting/republishing this material for advertising or promotional purposes, creating new collective works, for resale or redistribution to servers or lists, or reuse of any copyrighted component of this work in other works.}
\newcommand\copyrightnotice{
	\begin{tikzpicture}[remember picture,overlay]
	\node[anchor=south,yshift=10pt] at (current page.south) {\fbox{\parbox{\dimexpr\textwidth-\fboxsep-\fboxrule\relax}{\copyrighttext}}};
	\end{tikzpicture}}

\newif\ifIEEE  %

\title{\LARGE \bf Learning to Predict Ego-Vehicle Poses for Sampling-Based Nonholonomic Motion Planning}

\author{Holger Banzhaf$^1$, Paul Sanzenbacher$^2$, Ulrich Baumann$^2$, J. Marius Z\"ollner$^3$ %
	\thanks{$^{1}$Holger Banzhaf is with Robert Bosch GmbH, Corporate Research, Automated Driving, Renningen, Germany.}
	\thanks{$^{2}$Paul Sanzenbacher and Ulrich Baumann were with Robert Bosch GmbH, Corporate Research, Automated Driving, Renningen, Germany.}
	\thanks{$^{3}$J. Marius Z\"ollner is with FZI Research Center for Information Technology, Karlsruhe, Germany.} 
}

\begin{document}	
\maketitle
\ifIEEE
\else
	\copyrightnotice \vspace{-10pt}  %
\fi
\thispagestyle{empty}                                       %
\pagestyle{empty}

\setlength{\floatsep}{0.8\baselineskip plus  0.2\baselineskip minus  0.2\baselineskip} %
\setlength{\intextsep}{0.8\baselineskip plus 0.2\baselineskip minus  0.2\baselineskip} %
\setlength{\textfloatsep}{1.4\baselineskip plus  0.2\baselineskip minus  0.4\baselineskip} %

\begin{abstract}
Sampling-based motion planning is an effective tool to compute safe trajectories for automated vehicles in complex environments. However, a fast convergence to the optimal solution can only be ensured with the use of problem-specific sampling distributions. Due to the large variety of driving situations within the context of automated driving, it is very challenging to manually design such distributions. This paper introduces therefore a data-driven approach utilizing a deep convolutional neural network (CNN): Given the current driving situation, future ego-vehicle poses can be directly generated from the output of the CNN allowing to guide the motion planner efficiently towards the optimal solution. A benchmark highlights that the CNN predicts future vehicle poses with a higher accuracy compared to uniform sampling and a state-of-the-art A*-based approach. 
Combining this CNN-guided sampling with the motion planner Bidirectional RRT* reduces the computation time by up to an order of magnitude and yields a faster convergence to a lower cost as well as a success rate of \SI{100}{\percent} in the tested scenarios.
\end{abstract}

\section{Introduction} \label{sec:introduction}
\begin{figure}[t]
	\centering
	\includegraphics[width = 0.94\columnwidth]{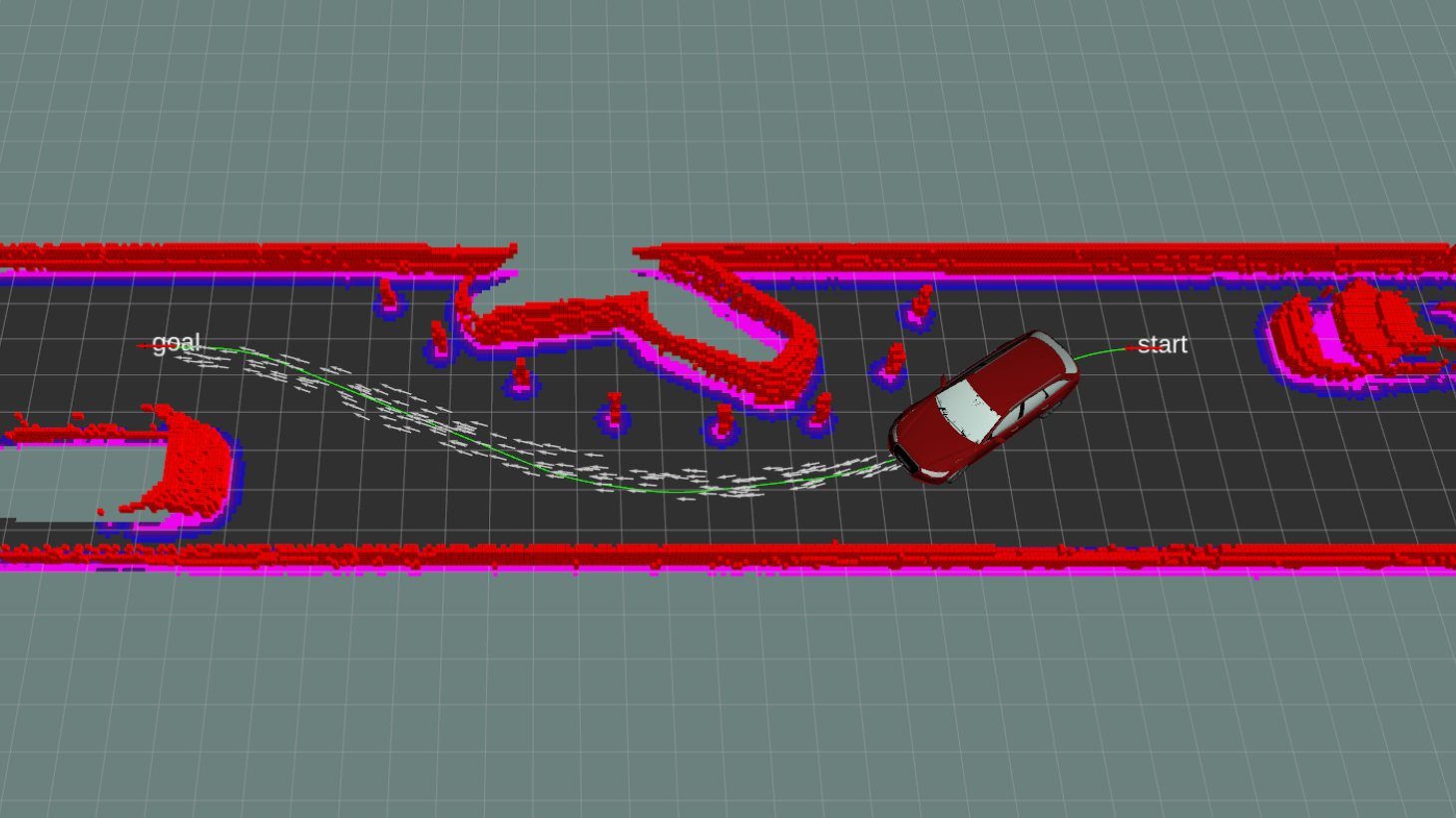}
	\caption{Evasive maneuver due to an accident blocking the road. The path of the ego-vehicle (green line) is computed using the learned vehicle pose predictions (gray arrows). The obstacles in the environment are visualized by the red voxels, and the corresponding two-dimensional cost map for motion planning is depicted on the ground.}
	\label{fig:title}
	\vspace{-0.25cm} %
\end{figure}

Motion planning is one of the major pillars in the software architecture of automated vehicles. Its task is to compute a safe trajectory from start to goal while taking into account the vehicle's constraints, the non-convex surrounding as well as the comfort requirements of passengers. In structured environments, such as highway driving, it is sufficient to solve the motion planning problem locally close to the lane centerline. Semi-structured environments or complex maneuvers, however, require a global solution as local approaches typically fail. Real-world examples of such scenarios are evasive maneuvers due to blocked roads (see \cref{fig:title}), multi-point turns at dead ends, or various parking problems.

Despite the complexity of such situations, the real-time constraints of automated vehicles still require a fast convergence to a preferably optimal solution. In order to achieve this, a common approach in the literature is the design of hand-crafted heuristics that guide the motion planner towards the goal~\cite{dolgov2010path, chen2015path}. However, finding a good heuristic that takes into account the nonholonomic constraints of the vehicle while adapting to the various driving situations is nearly as difficult as solving the motion planning problem itself.

An appealing solution for this dilemma is the combination of classic motion planning techniques~\cite{gonzalez2016review} with machine learning (ML) algorithms that learn to guide the motion planner in a data-driven way. In such a tandem configuration, for instance, the ML algorithm generates potential solutions for the planning problem while the motion planner selects the best proposals and guarantees collision avoidance.

This paper introduces and evaluates such a combination, in which a neural network guides the sampling-based motion planner Bidirectional RRT* (BiRRT*)~\cite{jordan2013optimal}. Sampling-based rather than search-based planning~\cite{dolgov2010path, likhachev2009planning} is considered here as the latter requires expert knowledge to discretize the state-action space. Especially in tight environments, an improper choice of the discretization resolution directly influences the completeness of the search-based planner. Sampling-based motion planners, on the other hand, require a problem-specific sampling distribution to speed up the planning process. Within this context, the main \textbf{contributions} are:
\begin{itemize}
	\item Prediction of future ego-vehicle poses given the current environment as well as the start and goal state using a deep convolutional neural network (CNN). The network is trained end-to-end using trajectories and observations recorded in a real-world driving simulator.
	\item Comparison of the proposed method with two baselines: uniform sampling and an A*-based approach. %
	\item Evaluation of the different sampling distributions along with the generic motion planner BiRRT* in three challenging automated driving scenarios. %
\end{itemize}

The remainder of this paper is organized as follows: \cref{sec:related_work} describes the related work, and \cref{sec:learning} focuses on the neural network used for the prediction task. The performance of the CNN-guided motion planner is highlighted in \cref{sec:guided_mp}, followed by a conclusion in \cref{sec:conc}. A supplementary video can be found at \url{https://youtu.be/FZPn3OHdQxk}. 

\section{Related Work} \label{sec:related_work}

\vspace{-13.4pt} %

Recent advances in deep learning have opened up new possibilities to improve or even replace existing motion planning algorithms for robot navigation. For instance, impressive results have been achieved in the field of imitation learning, where a robot is trained to act based on expert demonstrations, such as end-to-end learning for self-driving cars. Here, deep neural networks have shown the capability to learn reactive policies given high-dimensional sensor inputs~\cite{pomerleau1989alvinn,bojarski2016end}. However, these approaches cannot guarantee safety and might fail in situations where a global solution of the motion planning problem is required.

The shortcomings of pure ML-based solutions can be avoided by replacing only non-safety-relevant parts of existing motion planners with ML, and hence maintain crucial properties like safety and optimality. Depending on the planning algorithm, different approaches exist in the literature. Optimization-based planners, for example, often suffer from short planning horizons. This issue can be resolved with a learned cost shaping term that aligns the short-term horizon with the long-term goal as proposed in \cite{tamar2017learning}.

In contrast to that, search-based planners in continuous state-action space require an action sampling distribution for graph expansion and a cost-to-go heuristic for node selection. The former is addressed in \cite{kim2018guiding}, where a generative adversarial network \cite{goodfellow2014generative} is trained to guide the search towards the goal. Minimizing search effort through learned cost-to-go heuristics is achieved in \cite{choudhury2018data}. However, iteratively evaluating the heuristic during search limits the capacity of the neural network due to real-time constraints. Furthermore, optimality can only be guaranteed in a multi-heuristic setup.

Sampling-based motion planners, on the other hand, require problem-specific sampling distributions for fast convergence to the optimal solution. A promising research direction is to learn these distributions using imitation learning. For instance, \cite{hubschneider2017integrating} integrates end-to-end learning in a sampling-based motion planner. A conditional variational auto-encoder~\cite{sohn2015learning} is trained in \cite{ichter2018learning} to learn a sampling distribution of arbitrary dimension. While this approach seems generally promising, scaling it up to high-dimensional environment models, such as occupancy grids, remains an unresolved challenge. Opposed to that, \cite{perez2018learning} uses a CNN to predict future positions of the robot given a start and goal position as well as the current environment. The CNN's output is then used to bias the sampling of RRT* \cite{karaman2010incremental}. A similar idea is presented in this paper with the major difference that not only position, but entire pose predictions (position and orientation) are learned for nonholonomic motion planning.

Recent publications \cite{qureshi2018motion, ichter2018robot, srinivas2018universal} have shown novel motion planning algorithms that conduct planning in a learned latent space rather than in the complex configuration space. The general idea is to simplify the planning problem by solving it in the lower-dimensional latent space. While still in an early development phase, the future of these approaches in safety-critical applications highly depends on the possibility to satisfy hard constraints like collision avoidance.

\section{Learning Ego-Vehicle Pose Predictions} \label{sec:learning}

This section introduces the model for the prediction of the ego-vehicle poses connecting a start with a goal state in an environmental aware fashion. \cref{subsec:data,subsec:model,subsec:sampling} highlight the data generation process, the model, and the sampling of vehicle poses from the model's output. The training process, hyperparameter optimization, and evaluation are finally described in \cref{subsec:training,subsec:hyper_opti,subsec:evaluation}.

\subsection{Data Generation} \label{subsec:data}

Learning ego-vehicle predictions in a supervised fashion requires a diverse dataset consisting of the vehicle's trajectory from start to goal and the corresponding observations of the environment. Such a dataset with a total number of \num{13418}~trajectories is recorded in a real-world simulator using Gazebo and ROS. The implemented data generation process is fully automated and parallelized with multiple simulation instances, and requires no time-consuming manual labeling of the recorded training data.

\begin{figure*}[htbp]
	\vspace{0.25cm} %
	\centering
	\ifIEEE
\else
		\includegraphics[width = 1.95\columnwidth]{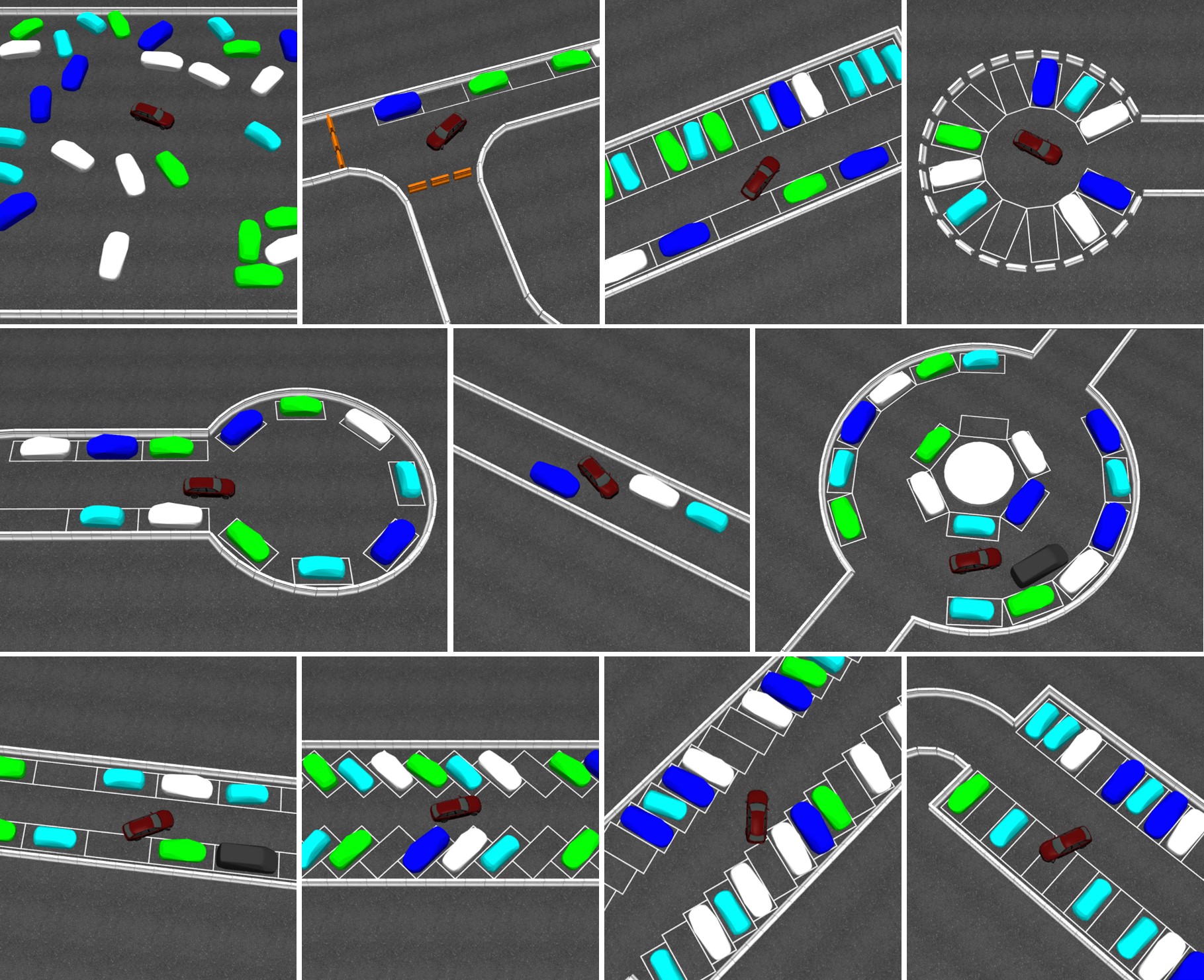}
		\caption{Visualization of the eleven scenarios used in the generation of the dataset. Each scenario highlights a challenging driving task in the context of low-speed maneuvering in cluttered environments. First row: arena with a maze-like structure, blocked T-intersection, urban parking environment, and circular parking lot. Second row: dead end, blocked road, and cluttered roundabout. Third row: various parking scenarios, where the angle between the driveway and the parking spot is varied between \SI{0}{\degree}, \SI{45}{\degree}, \SI{75}{\degree}, and \SI{90}{\degree}.}
	\fi
	\label{fig:scenarios_train}
	\vspace{-0.25cm} %
\end{figure*}

Each recording is generated in one of the eleven scenarios visualized in \cref{fig:scenarios_train}.
As this research is embedded in a project with a focus on low-speed maneuvering in tight environments, the designed scenarios focus on challenging setups that require the vehicle to act precisely in a highly non-convex workspace. Exemplary situations include blocked roads, dead ends, or different parking problems. Variation is introduced in each scenario by changing the start and goal pose of the ego-vehicle as well as the number and location of the static obstacles. For unstructured environments, this is implemented by randomly sampling the obstacles and the vehicle's start and goal pose. For semi-structured environments, obstacles are randomly assigned to predefined locations, such as parking spots. The ego-vehicle's start and goal pose are, in this case, randomly chosen within predefined locations corresponding to the entrance and exit of a scenario, the driveway, or a parking spot.

The following procedure is then applied to obtain a recording in the randomly configured instances of each scenario. First, the motion planner BiRRT* generates a curvature-continuous collision-free path from start to goal as detailed in \cref{sec:guided_mp}. Here, BiRRT* is guided by the A*-based heuristic described in \cref{subsec:evaluation}, and its initial motion plan is optimized for \SI{5}{s}. Next, a motion controller executes the computed solution resulting in a trajectory with $t=1, \ldots, T$ vehicle states $\state_t \in \states$, where a state $\state_t = (x_t, y_t, \theta_t, \kappa_t, v_t)\transpose$ at time step $t$ is defined by the vehicle's position $(x_t, y_t)$, orientation $\theta_t$, curvature $\kappa_t$, and velocity $v_t$. Note that throughout this paper, the term vehicle pose is only used to describe a lower-dimensional subset of the state including the vehicle's position and orientation. Finally, the observations from a simulated LiDAR perception are fused in a two-dimensional \SI{60 x 60}{\meter} occupancy grid with a resolution of \SI{10}{\centi\meter} and recorded along the trajectory. A visualization of such a recording can be found in \cref{fig:cnn_inputs} on the left. 

It has to be noted that, similar to \cite{choudhury2018data}, motion planning for data generation is conducted with full environmental information in order to guarantee fast convergence to a cost-minimizing solution. In contrast to that, the recorded occupancy grid only fuses the current history of sensor observations resulting in unobserved areas due to occlusions. This forces the model in the learning phase to also resolve situations with partial observability that requires an intuition where possible solutions might lie.

\subsection{Model} \label{subsec:model}

The model is designed to generate a sampling distribution over future vehicle poses connecting a start and goal state given an occupancy grid with environmental observations. Previous publications~\cite{perez2018learning, baumann2018path, caltagirone2017lidar} have shown that CNNs are well suited for processing high dimensional inputs and predicting multi-modal distributions over future ego-vehicle locations. For nonholonomic motion planning, however, it is essential to enrich such a spatial distribution with the information about the robot's heading angle at all predicted positions. Therefore, it is proposed to jointly learn a sampling distribution over future vehicle positions as well as a mapping from position to heading angle for a complete representation of a vehicle pose.

\begin{figure*}[htbp]
	\vspace{0.25cm} %
	\centering
	\includegraphics[width = 1.95\columnwidth]{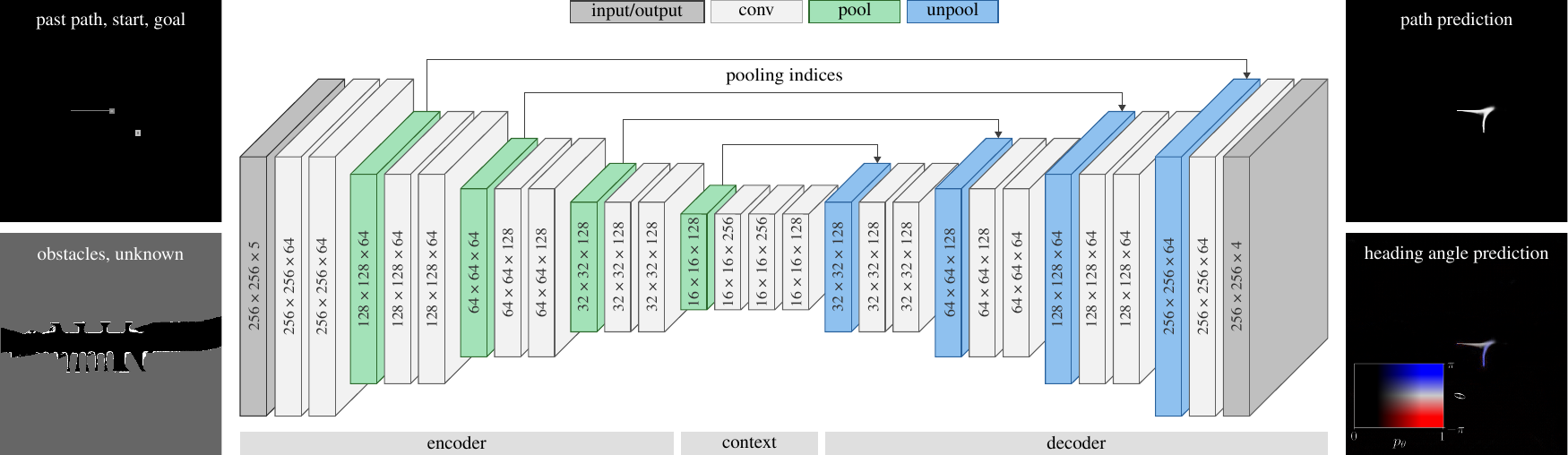}
	\caption{Architecture of the CNN, which predicts a two-dimensional sampling distribution over future ego-vehicle positions (top right) and the corresponding mapping from position to heading angle (bottom right). Note that the five input girds, which describe the current driving situation, are blended into two images on the left.}
	\label{fig:cnn_architecture}
\end{figure*}

This task is realized with the encoder-decoder architecture shown in \cref{fig:cnn_architecture}. The illustrated CNN, which contains about \num{2.5}~million parameters, is based on the well-known \mbox{SegNet}~\cite{badrinarayanan2015segnet} from semantic segmentation. The main contribution is not the architecture of the model itself, as it can be easily exchanged with any other state-of-the-art architecture, but the representation of the input and output layers.

\begin{figure*}[htbp]
	\centering
	\includegraphics[width = 1.95\columnwidth]{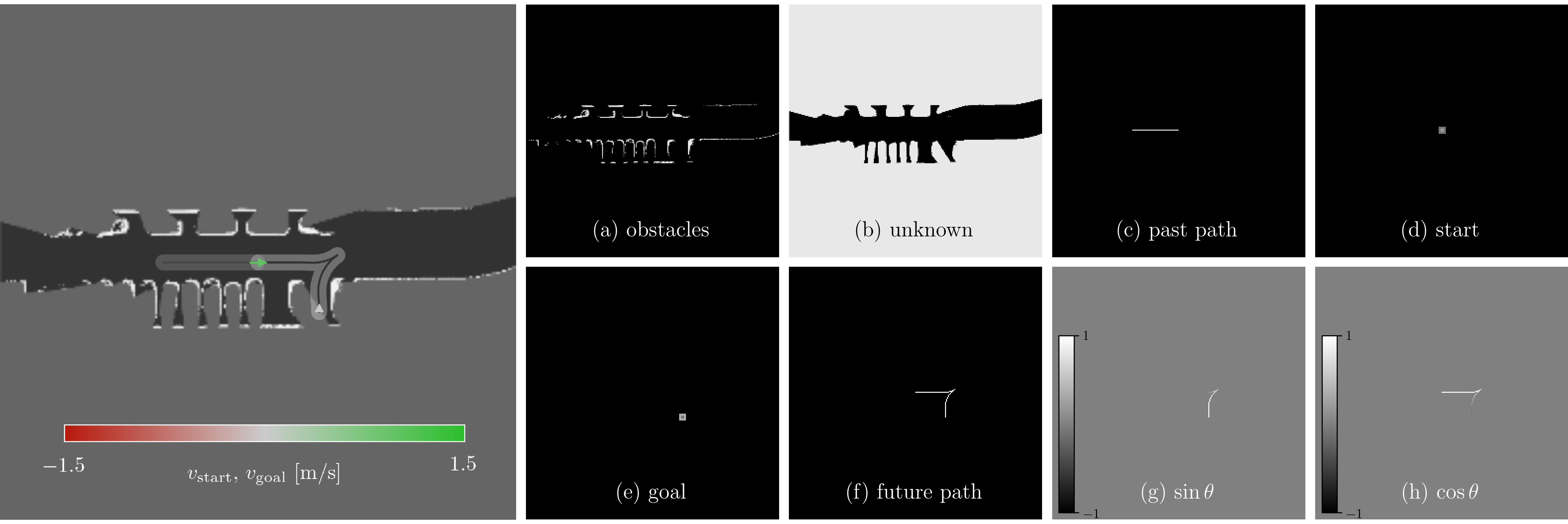}
	\caption{Visualization of a perpendicular parking task on the left, and the derived input grids (a)--(e) and label grids (f)--(h) for training the CNN. The ego-vehicle is represented by the green arrow in the center of the image on the left and the target vehicle pose by the white arrow without a tail. The color of the arrows indicates the vehicle's velocity at the respective pose.}
	\label{fig:cnn_inputs}
	\vspace{-0.25cm} %
\end{figure*}

The proposed network takes five grids with a resolution of \SI{256 x 256}{\pixel} as an input. These grids encode the static obstacles, the unknown environment, the past path, and the start and goal state of the vehicle (see \cref{fig:cnn_inputs}).
A simple and effective encoding of the vehicle state is achieved by a \SI{7 x 7}{\pixel} square that describes three information: (1) its location in the grid marks the position of the vehicle, (2) its inner pixels encode the corresponding vehicle velocity, and (3) its outer pixels depict the respective heading angle. Note that the curvature of the start state is implicitly encoded in the past path as it can be seen in \cref{fig:cnn_architecture} at the top left.

The input of the CNN is first encoded and then decoded back to its original size using alternating blocks of convolutional layers with \num{3x3}~kernels and a stride of \num{1}, \num{2x2}~max pooling layers, and \num{2x2}~unpooling layers. All convolutional layers except the last one are followed by a batch normalization~\cite{ioffe2015batch} and a ReLU activation~\cite{nair2010rectified}.

The CNN outputs four grids of the same resolution as the input grids. The first two grids give the results of a cell-wise classification that is trained to predict whether a given cell belongs to the future path or not. An example of such a prediction is shown in \cref{fig:cnn_architecture} at the top right, where the intensity corresponds to the probability $\probPath(c)$ of a cell~$c$ being part of the future path. In contrast to that, the last two output grids contain the results of a cell-wise regression for the sine and cosine components of the robot's heading angle. The decomposition into sine and cosine has three advantages: (1) a cell with zeros in both components represents an invalid orientation and can therefore be used to label cells without angle information (see \cref{fig:cnn_inputs}), (2) computing the cell-wise norm of the predicted components can be interpreted as a confidence measure $p_{\theta}(c)$ indicating if an angle information is available at a respective cell~$c$ (see \cref{fig:cnn_architecture} at the bottom right), and (3) the prediction of the heading angle can be treated as a regression  task rather than a classification task, which yields a compact representation of the output and avoids an exponential growth of its dimension. However, the latter comes with a potential drawback of not being able to predict multi-modal, cell-wise heading angle predictions. The experiments have shown, however, that this is only an issue in rare corner cases as most of the scenarios do not require the vehicle to change its heading angle significantly while staying in the same region of the environment.

\subsection{Vehicle Pose Sampling} \label{subsec:sampling}

Generating $i=1, \ldots, N$ continuous vehicle poses $\statePred^{[i]}$ from the output of the CNN is conducted in four steps. First, $N$ random cells $\cellPred^{[i]}$ are sampled from the CNN's path prediction $\probPath(c)$, which can be seen as a probability mass function (pmf) that describes the relative probability of a cell~$c$ being part of the future path (see \cref{fig:cnn_architecture}). Sampling from the pmf can be realized efficiently using the low-variance sampling algorithm in~\cite{thrun2005probabilistic}. Its linear time complexity yields a fast computation even for large values of $N$. Next, a continuous position prediction $(\xPred^{[i]},\yPred^{[i]})$ is obtained by sampling uniformly within the previously computed discrete cell $\cellPred^{[i]}$. The heading angle prediction $\thetaPred^{[i]}$ is now evaluated at the cell $\cellPred^{[i]}$. If required, it can be interpolated with respect to the continuous position $(\xPred^{[i]},\yPred^{[i]})$, which is omitted here for the sake of simplicity. In a final step, the sampled position and the corresponding heading angle are concatenated yielding a sampled vehicle pose $\statePred^{[i]}$.

Note that throughout this paper, samples are exclusively drawn from cells with $\probPath(c) > \num{0.5}$ in order to only guide the motion planner towards regions with a high probability to be part of the future path. In contrast to that, visualizations of $\probPath(c)$ are never thresholded and show the unmodified output of the CNN.

\subsection{Training and Metrics} \label{subsec:training}

The proposed model is trained end-to-end using \SI{64}{\percent} of the recorded trajectories. The training dataset is further augmented by randomly selecting up to \num{100} different start states on a recorded trajectory resulting in \num{807273} (partly correlated) data points. This augmentation strategy does not only upscale the dataset, but also forces the CNN to adjust its prediction as more information on the vehicle's environment becomes available.

The parameters of the network are optimized using Adam~\cite{kingma2015adam} and the loss function
\begin{equation}
	L = \mathsmaller{\sum\limits_{c \in \cells}} \left( \fCE(c) \LCE(c) + \fMSE(c) \LMSE(c) \right) + \mathsmaller{\sum\limits_{w \in \weights}} \tfrac{\lambda w^2}{2},
\end{equation}
where the first term computes the cell-wise cross-entropy loss $\LCE(c)$ of the classification task, the second term the cell-wise mean squared error $\LMSE(c)$ of the regression task, and the last term the L2 regularization loss of the weights $w \in \weights$ with a scaling factor $\lambda$. As the majority of cells in the label grids contain no path information (see \cref{fig:cnn_inputs}), a weighting of the relevant cells is conducted with the functions $\fCE(c)$ and $\fMSE(c)$ given as
\begin{align}
	\fCE(c) &= 1 + \indicator_c \cdot (\gammaCE - 1), \\
	\fMSE(c) &= 1 + \indicator_c \cdot (\gammaMSE -1),
\end{align}
where $\gammaCE$ and $\gammaMSE$ are hyperparameters of the model, and $\indicator_c$ is the indicator function that is \num{1} if the future path crosses a cell $c$ and \num{0} otherwise.

In order to evaluate the prediction capability of the model, two metrics are introduced below. The first metric measures the proximity of the $N$ predicted samples $\statePred^{[i]} \in \statesPred$ to the ground truth trajectory $\state_t \in \states$ by computing the average path deviation~$D$ according to
\begin{align}
	D(\states, \statesPred) = \frac{\sum_{i=1}^{N} \min_{\state_t \in \states} d(\state_t, \statePred^{[i]})}{N},
\end{align}
where $d(\bullet)$ describes the distance between a pose on the trajectory and a sample. It is computed by a weighted sum of the Euclidean distance and the angular deviation given as
\begin{align}
	d(\state_t, \statePred^{[i]}) = w_{\text{pos}} \norm{ \statePred^{[i]} - \state_t}_2 + w_{\theta}\left| \thetaPred^{[i]} - \theta_t \right|, \label{eq:weighted_distance}
\end{align}
where the different units of both terms are taken into account by setting $w_\text{pos}$ to \num{0.35} and $w_{\theta}$ to \num{0.65} throughout this paper.

As the predicted poses are supposed to guide the motion planner through complex scenarios, it is beneficial to have evenly distributed samples along the ground truth trajectory. Therefore, the second metric measures the maximum prediction gap $G$ in the following two steps: (1) project every sample onto the trajectory, and (2) evaluate the maximum arc length that is not covered by any sample. The projection in step~1 is defined as
\begin{align}
	\stateRef^{[i]} &= \argmin_{\state_t \in \states} d(\state_t, \statePred^{[i]}),
\end{align}
where $\stateRef^{[i]}$ is the reference pose of the predicted sample $\statePred^{[i]}$ on the trajectory. These reference poses are then added to the ordered list $\statesRef$ according to their arc length $s_\text{ref}^{[i]}$. The maximum prediction gap $G \in [0,1]$ is finally obtained by
\begin{align}
	G(\states, \statesRef) = \frac{\max_{i=1:N-1} s_\text{ref}^{[i+1]} - s_\text{ref}^{[i]}}{s_T},
\end{align}
where the length of the recorded trajectory is denoted by $s_T$.

\subsection{Hyperparameter Optimization} \label{subsec:hyper_opti}

\ifIEEE
\else

This subsection evaluates the influence of the hyperparameters on the CNN's prediction performance based on the previously derived metrics. The analysis is conducted on a validation dataset containing \SI{16}{\percent} of the recorded trajectories. Similar to the training dataset, up to five different start states are selected on each trajectory resulting in \num{10730} (partly correlated) data points.

The different parametrizations of the model are trained on an Nvidia Titan~X with a batch size of \num{20} using the Python interface of TensorFlow. Training a model to convergence takes about \SI{30}{\hour}, which corresponds to \num{90000} iterations. A visualization of the training statistics for one of the conducted optimizations can be found in \cref{fig:training_stats}.
\begin{figure}[htbp]
	\centering
	\includegraphics[width = 0.95\columnwidth]{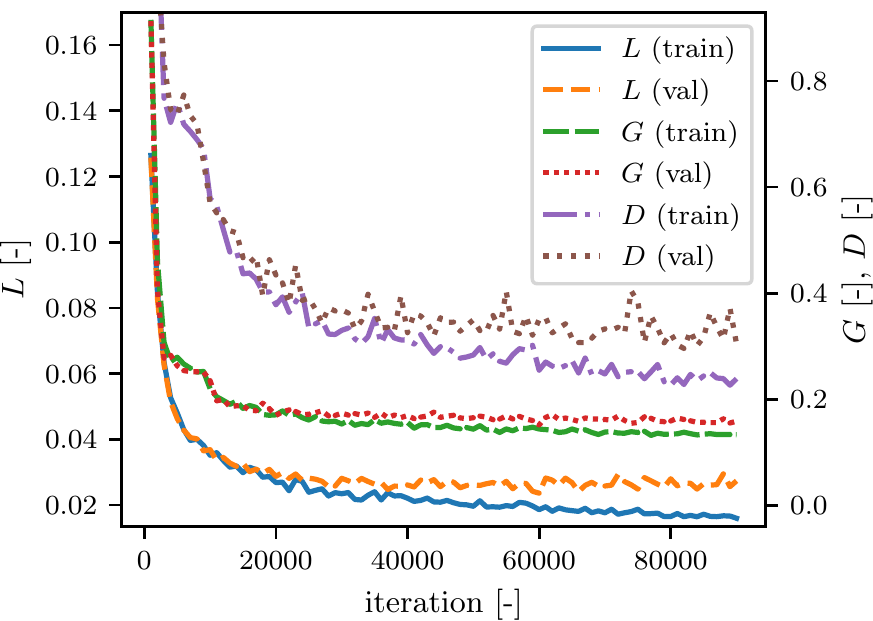}
	\caption{Visualization of the training statistics with the hyperparameters set to $\gammaCE=\gammaMSE=25$ and $lr=10^{-5}$ (learning rate). Note that for better comparability, the loss $L$ is visualized without the L2 regularization term as it is only computed at training and not at inference time. The metrics $G$ and $D$ are evaluated on \num{50}~sampled vehicle poses.}
	\label{fig:training_stats}
\end{figure}

\begin{figure*}[htbp]
	\vspace{0.25cm} %
	\centering
	\begin{subfigure}{0.312\textwidth}
		\centering	
		\includegraphics[width=1.0\textwidth]{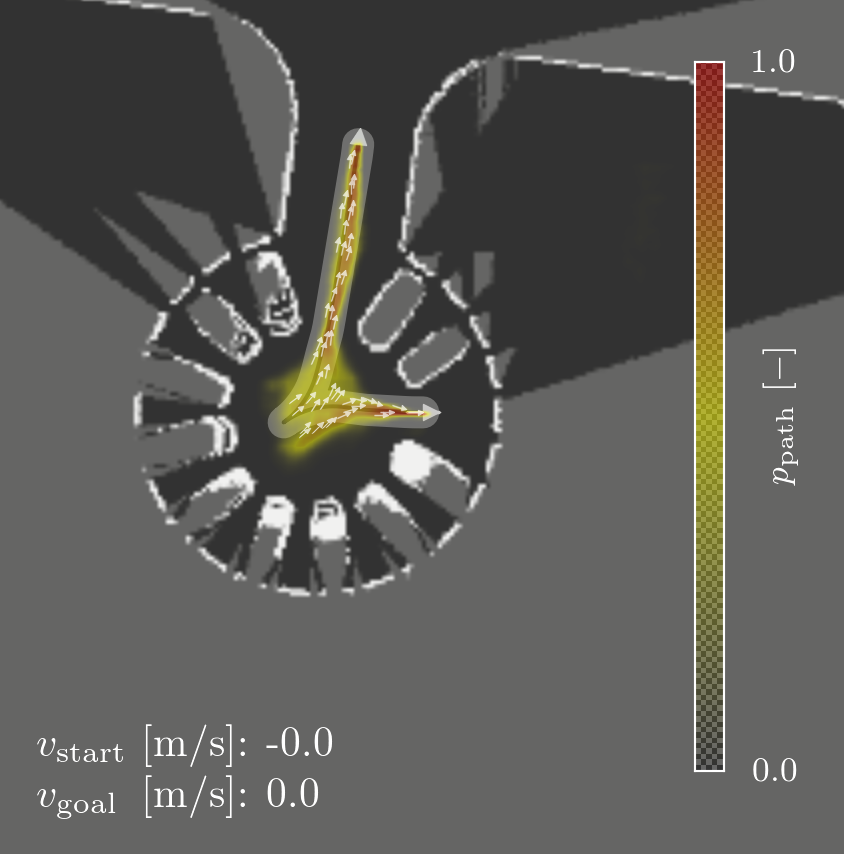}
		\subcaption{$lr = \num{e-5}$, $\gammaCE = \num{10}$, $\gammaMSE = \num{100}$}
	\end{subfigure}
	\begin{subfigure}{0.312\textwidth}
		\centering
		\includegraphics[width=1.0\textwidth]{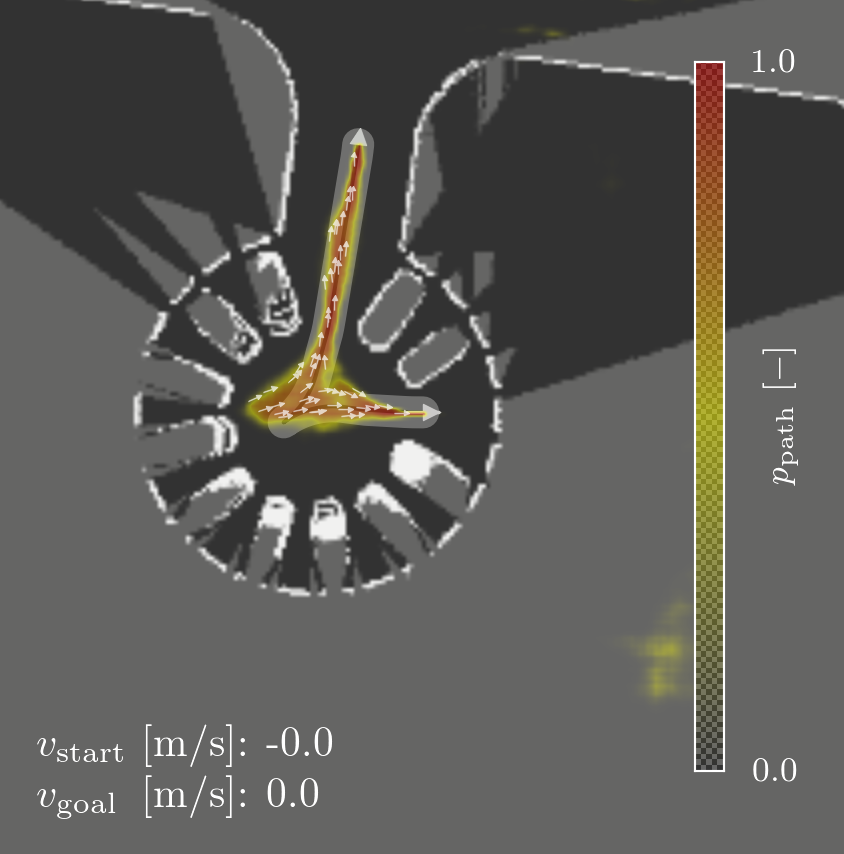}
		\subcaption{$lr = \num{e-5}$, $\gammaCE = \num{25}$, $\gammaMSE = \num{25}$}
	\end{subfigure}
	\begin{subfigure}{0.312\textwidth}
		\centering
		\includegraphics[width=1.0\textwidth]{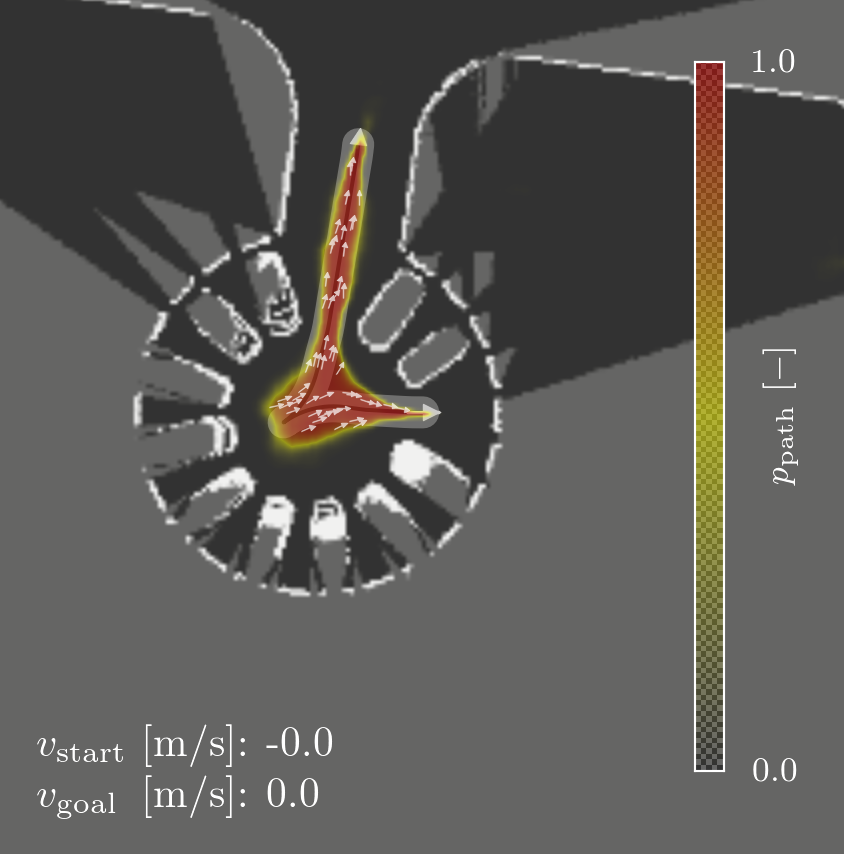}
		\subcaption{$lr = \num{e-5}$, $\gammaCE = \num{100}$, $\gammaMSE = \num{10}$}
	\end{subfigure}
	\caption{Qualitative comparison of the different hyperparameters on a test set trajectory with \num{50}~sampled vehicle poses.}
	\label{fig:hyper_opti}
	\vspace{-0.25cm} %
\end{figure*}

In order to decrease the exponentially growing search effort, only the hyperparameters with the highest expected influence on the CNN's overall performance are optimized. Therefore, two out of the five hyperparameters, namely the exponential learning rate decay $lrd$ and the L2 regularization scaling factor $\lambda$, are fixed to $lrd=\num{0.01}$ after \num{e6} iterations and $\lambda=\num{0.003}$. The remaining hyperparameters are optimized in two steps. First, the learning rate $lr$ is varied while keeping $\gammaCE$ and $\gammaMSE$ at~\num{25}, and second, $lr$ is set to its best value while $\gammaCE$ and $\gammaMSE$ are optimized.

The quantitative results of the hyperparameter optimization are listed in \cref{tab:hyper_opti}.
\begin{table}[htbp]
	\caption{Quantitative comparison of the hyperparameter optimization on the basis of \num{200} sampled vehicle poses carried out on the validation dataset. The displayed values are given with mean and standard deviation.}
	\label{tab:hyper_opti}
	\centering
	\begin{tabular}{ccccc}
		\toprule
		$lr$ & $\gammaCE$ & $\gammaMSE$ & $G$\,[\si{\percent}] &  $D$\,[\si{-}] \\
		\midrule
		$10^{-3}$ & \multirow{3}{*}{25} & \multirow{3}{*}{25} & 20.7 $\pm$ 18.9 & 0.64 $\pm$ 0.26 \\
		$10^{-4}$ & & & 16.9 $\pm$ 18.6 & 0.35 $\pm$ 0.33 \\
		$10^{-5}$ & & & 10.1 $\pm$ 12.0 & 0.33 $\pm$ 0.35 \\
		\midrule
		\multirow{8}{*}{$10^{-5}$} & 10 & 10 & 12.6 $\pm$ 14.9 & 0.26 $\pm$ 0.26 \\
		& 10 & 25 & 12.7 $\pm$ 15.3 & 0.26 $\pm$ 0.28 \\
		& 10 & 100 & 12.5 $\pm$ 14.9 & 0.25 $\pm$ 0.25  \\
		& 25 & 10 & 11.2 $\pm$ 13.4 & 0.31 $\pm$ 0.35 \\
		& \textbf{25} & \textbf{25} & \textbf{10.1 $\pm$ 12.0} & \textbf{0.33 $\pm$ 0.35} \\
		& 25 & 100 & 10.8 $\pm$ 12.9 & 0.33 $\pm$ 0.33 \\
		& 100 & 10 & \phantom{0}9.8 $\pm$ 10.6 & 0.45 $\pm$ 0.40 \\
		& 100 & 25 & 10.3 $\pm$ 11.6 & 0.46 $\pm$ 0.45 \\
		& 100 & 100 & 10.3 $\pm$ 11.5 & 0.42 $\pm$ 0.35 \\
		\bottomrule
	\end{tabular}
\end{table}
It can be seen that lower learning rates result in a better prediction performance as they decrease the maximum prediction gap metric~$G$ as well as the average path deviation metric~$D$. The insights from \cite{baumann2018path} that higher values for $\gammaCE$ incentivize the CNN to classify more cells as part of the future path can also be observed in \cref{tab:hyper_opti} and \cref{fig:hyper_opti}. Thus, increasing $\gammaCE$ reduces the maximum prediction gap~$G$ while raising the average path deviation~$D$ because more samples are placed further away from the ground truth. A potential risk of too large values for $\gammaCE$ is a deterioration of the heading angle prediction, which cannot be recovered by increasing $\gammaMSE$ (see \cref{tab:hyper_opti}). For the remaining evaluations, the model with $\gammaCE=\num{25}$, $\gammaMSE=\num{25}$, and $lr=\num{e-5}$ is selected as this combination yields a smooth distribution of the samples in close vicinity of the ground truth. 
\fi

\subsection{Evaluation} \label{subsec:evaluation}

The following paragraphs analyze the prediction capability of the CNN and benchmark its performance against uniform sampling and the A*-based approach called Orientation-Aware Space Exploration~(OSE)~\cite{chen2015path}. To the best of the authors' knowledge, the latter is currently one of the most effective approaches for guiding a motion planner through complex environments. It is based on an A* graph search and explores the workspace with oriented circles\ifIEEE\else~as illustrated in \cref{fig:ose}\fi.
\ifIEEE
\else
\begin{figure}[htbp]
	\centering
	\includegraphics[width = 0.95\columnwidth]{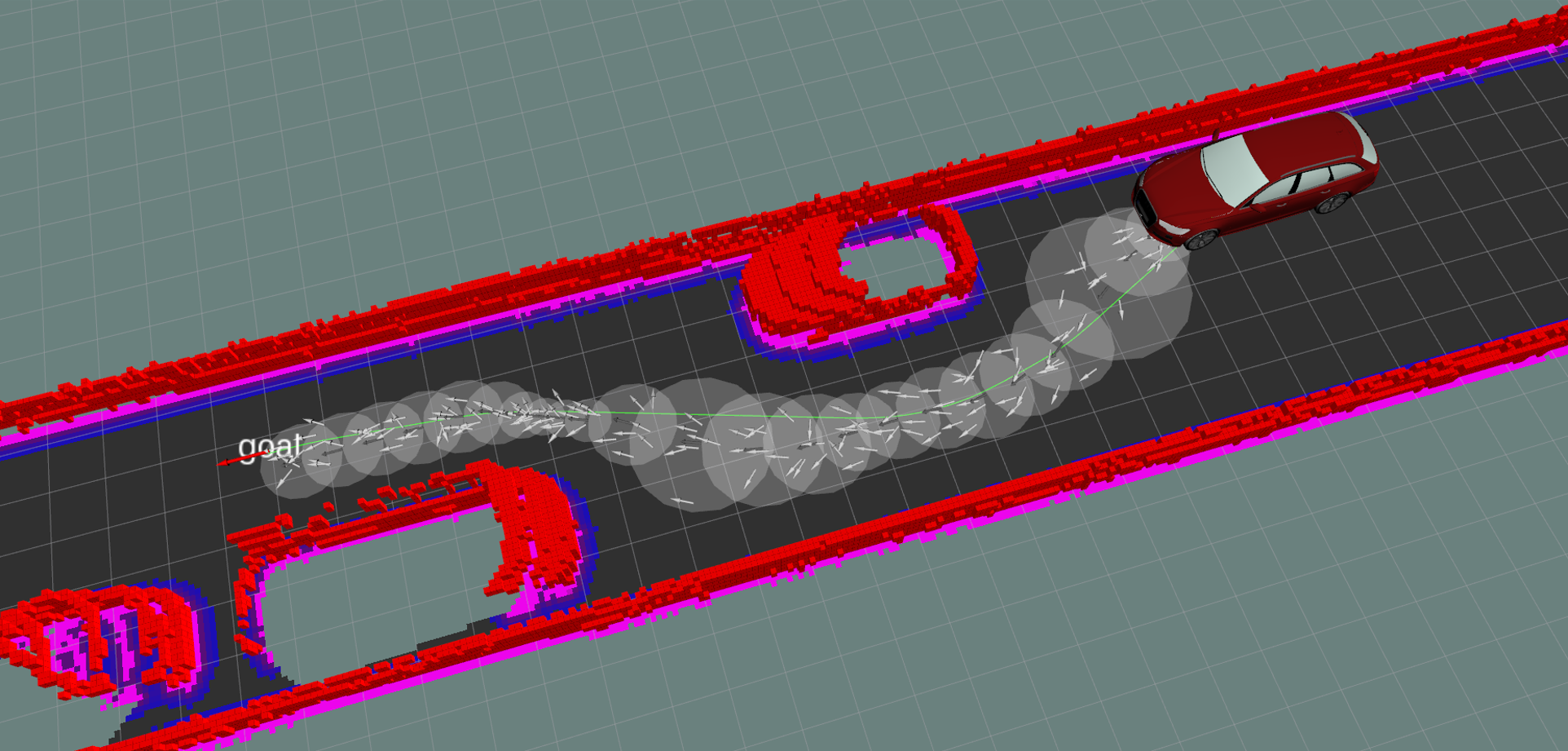}
	\caption{Orientation-Aware Space Exploration visualized by the white circles on the ground including \num{200}~sampled vehicle poses (gray arrows).}
	\label{fig:ose}
\end{figure}
\fi
Discrete vehicle poses can then be obtained by sampling from three-dimensional Gaussian distributions located at the center of the circles. \ifIEEE\else The standard deviation of the Gaussians can be made dependent of the radius~$r$ and is set to $\sigma_x = \sigma_y = r/3$ for the translational and $\sigma_{\theta} = \pi/6$ for the rotational component, respectively. Additional parameters of the OSE heuristic used in this paper are listed in \cref{tab:param_OSE}\fi.
\ifIEEE
\else
\begin{table}[htbp]
	\caption{Parameters of the Orientation-Aware Space Exploration.}
	\label{tab:param_OSE}
	\centering
	\begin{tabular}{ll}
		\toprule
		parameter & value \\
		\midrule
		minimum radius & \SI{0.2}{\metre} \\ 
		maximum radius & \SI{5.0}{\metre} \\
		minimum clearance & \SI{1.041}{\metre} \\
		neighbors & \num{32} \\
		maximum curvature & \SI{0.1982}{\per\metre} \\
		computation timeout & \SI{1.0}{\second} \\		
		\bottomrule
	\end{tabular}
\end{table}
\fi

\ifIEEE
\else
\begin{figure*}[htbp]
	\vspace{0.25cm} %
	\centering	
	\includegraphics[width = 1.95\columnwidth]{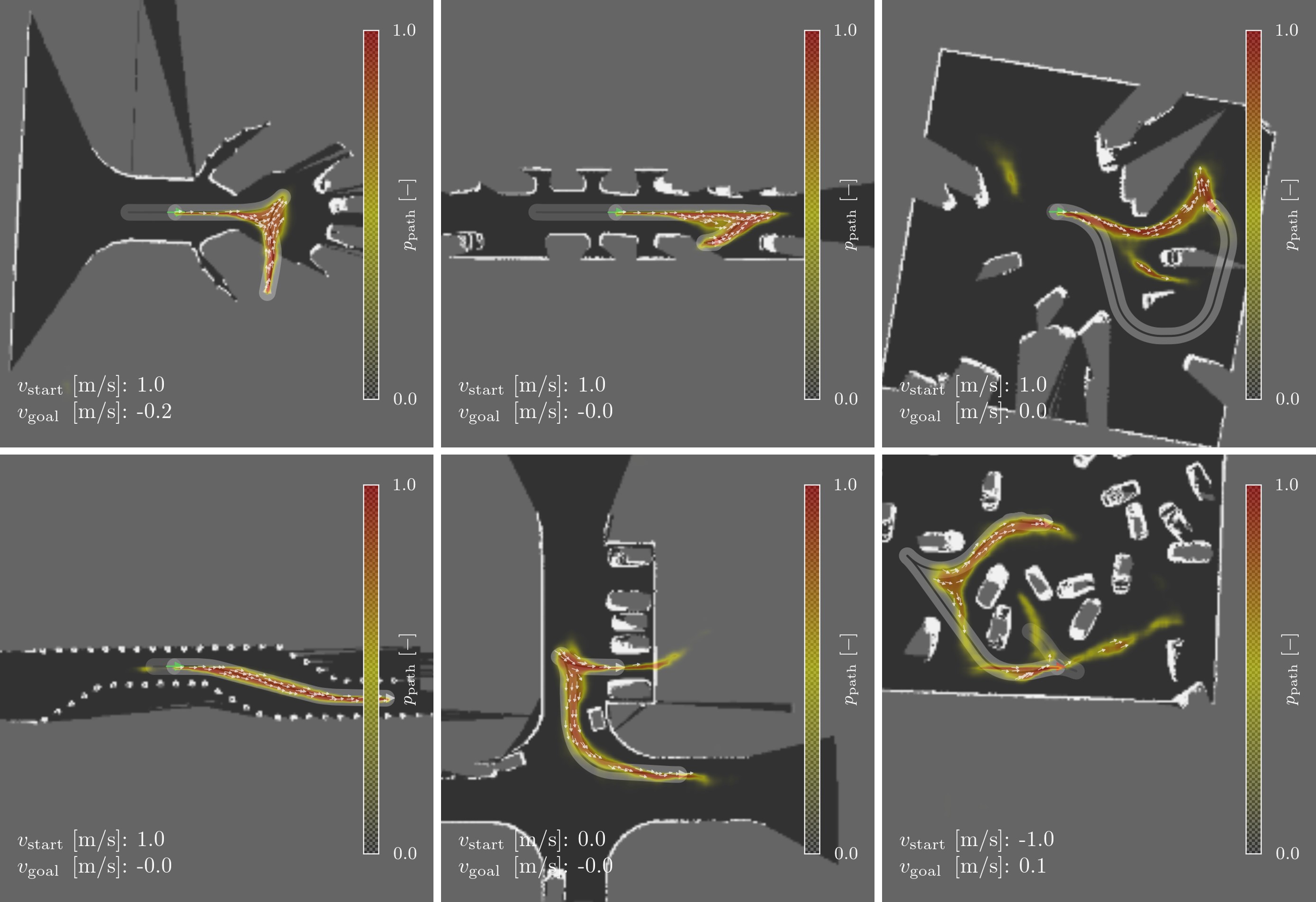}
	\caption{Illustration of the CNN prediction including \num{50}~sampled vehicle poses in different scenarios. The top row visualizes the performance on three test set trajectories recorded in the scenarios circular parking lot (left), parallel parking (middle), and arena (right). The bottom row displays the robustness of the approach in the novel scenarios construction zone (left) and cluttered 4-way intersection (middle). Both scenarios contain novel artifacts like construction barrels and dumpsters. The image on the bottom right shows a performance degeneration for a long prediction horizon in an unstructured environment.}
	\label{fig:cnn_predictions_all}
	\vspace{-0.25cm} %
\end{figure*}
\fi

A benchmark of the three approaches on the test dataset can be found in \cref{tab:comparison_heuristics}. The \num{13420}~test cases with up to five different start states on each trajectory represent the remaining \SI{20}{\percent} of the recordings. For a scenario-specific evaluation, \num{5770}~trajectories are extracted from this dataset, half of which come from the scenario arena and the other half from parallel and perpendicular parking. Both uniform sampling as well as the OSE procedure are evaluated on a single core of an Intel Xeon E5@\SI{3.5}{\giga\hertz} with the latter being implemented in C++. As opposed to that, the CNN is executed with its Python pipeline on an Nvidia Titan~X and an Intel Xeon E5@\SI{3.1}{\giga\hertz}.
\begin{table}[htbp]
\ifIEEE
\else
	\vspace{0.15cm} %
\fi
	\caption{Benchmark of uniform sampling, the OSE approach, and the CNN prediction on the test dataset. The metrics $G$ and $D$ as well as the computation time are evaluated on \num{200}~sampled vehicle poses and are given with mean and standard deviation.}
	\label{tab:comparison_heuristics}
	\centering
	\begin{tabular}{lcccc}
		\toprule
	 	& scenarios & $G$\,[\si{\percent}] & $D$\,[\si{-}] & time\,[\si{\milli\second}] \\
		\midrule
		uniform & all & 13.5 $\pm$ \makebox[0pt][l]{6.4}\phantom{00.0} & 7.84 $\pm$ 0.44 & 0.1 $\pm$ 0.0 \\
		OSE & all & 15.2 $\pm$ 17.4 & 0.57 $\pm$ 0.30 & 39.7 $\pm$ 43.1  \\
		CNN$^1$ & all  & 10.4 $\pm$ 12.5 & 0.34 $\pm$ 0.39 & 41.7 $\pm$ 14.0 \\
		CNN$^2$ & all & 11.8 $\pm$ 14.1 &  0.35 $\pm$ 0.40  & 39.7 $\pm$ 14.5 \\
		CNN$^3$ & all & 11.9 $\pm$ 14.3 &  0.34 $\pm$ 0.37  & 37.3 $\pm$ 13.7 \\
		CNN$^4$ & all & 17.8 $\pm$ 19.6  & 0.44 $\pm$ 0.35 & 37.4 $\pm$ 14.0 \\
		CNN$^5$ & all &  11.4 $\pm$ 14.0  & 0.32 $\pm$ 0.33  & 35.7 $\pm$ 14.5 \\
		CNN$^6$ & all &  12.0 $\pm$ 14.8  & 0.32 $\pm$ 0.33 & 38.8 $\pm$ 14.1 \\
		CNN$^7$ & all & 32.6 $\pm$ 22.4 & 0.82 $\pm$ 0.55  & 38.4 $\pm$ 13.4 \\		
		CNN$^1$ & arena & 14.4 $\pm$ 16.2 & 0.50 $\pm$ 0.47 & 42.5 $\pm$ 25.2 \\
		CNN$^1$ & parking & 9.1 $\pm$ 9.6 & 0.23 $\pm$ 0.15  & 40.5 $\pm$ 25.1 \\
		\bottomrule \addlinespace[1ex]
	\end{tabular}
	\begin{tabular}{ll}
		$^1$all input grids & $^2$all but the obstacle grid \\
		$^3$all but the unknown grid & $^4$all but the obstacle \& unknown grid \\
		$^5$all but the past path grid & $^6$all but the start grid \\
		$^7$all but the goal grid
	\end{tabular}
\ifIEEE
\else
	\vspace{-0.6cm} %
\fi
\end{table}

\ifIEEE
\else
\begin{figure*}[htbp]
	\vspace{0.25cm} %
	\centering
	\includegraphics[width = 1.95\columnwidth]{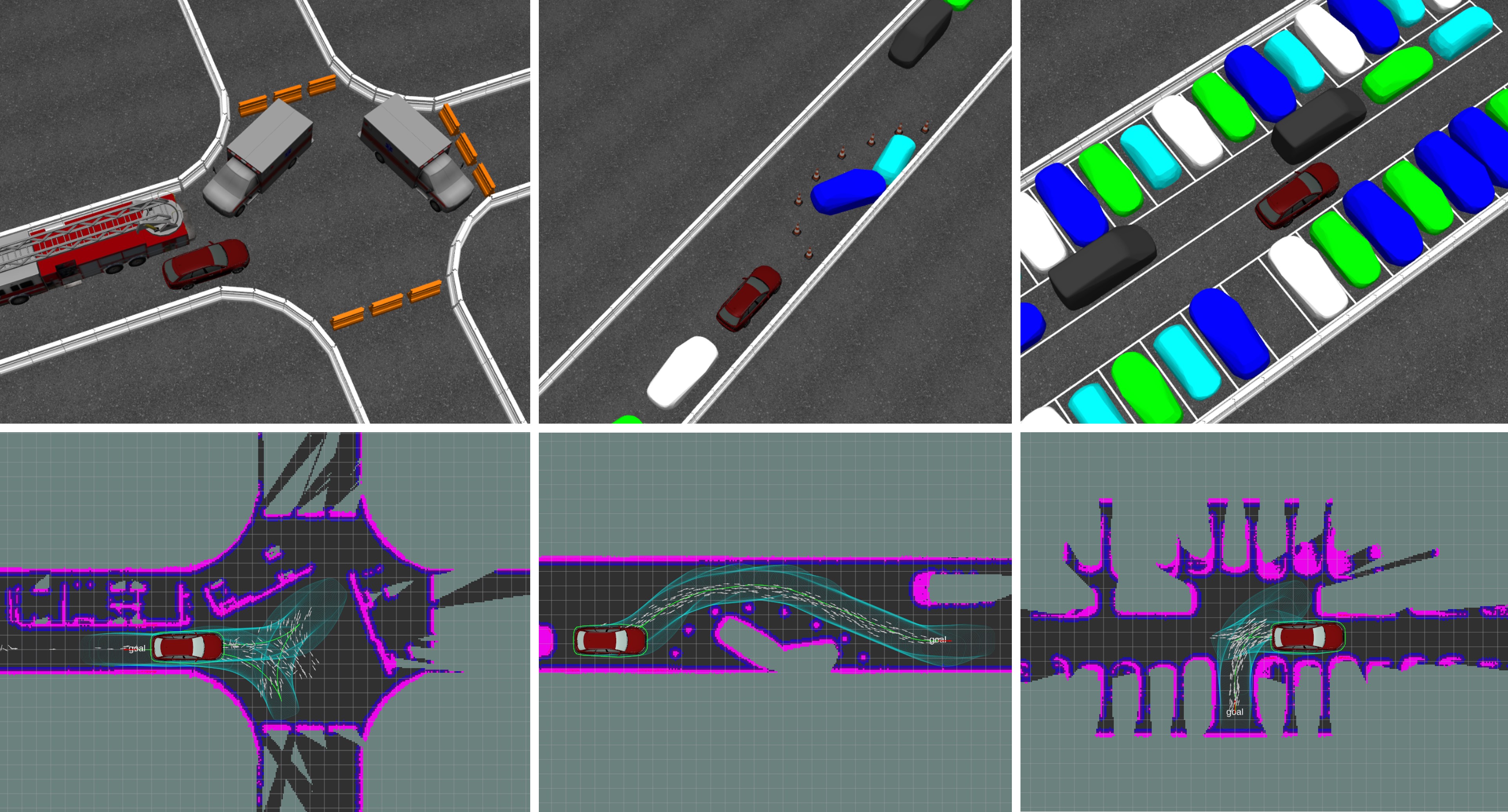}
	\caption{Guided motion planning in three automated driving scenarios. Exemplary solutions including \num{200} CNN-predicted vehicle poses (gray arrows) are shown in the bottom row. Scenario~I (left) illustrates a blocked intersection, where the red ego-vehicle has to execute a multi-point turn. A narrow passage problem due to a blocked road can be seen in Scenario~II (middle) while a high density parking environment~\cite{banzhaf2017deques} is displayed in Scenario~III (right).}
	\label{fig:scenarios_test}
	\vspace{-0.25cm} %
\end{figure*}
\fi

In comparison to uniform sampling and the OSE approach, \cref{tab:comparison_heuristics} shows that the CNN$^1$ predicts the vehicle poses more evenly along the ground truth path and yields overall smaller deviation from it. The mean computation time of the CNN is comparable to the OSE heuristic, however, with a one-third smaller standard deviation. The OSE's high variance in computation time is due to the fact that the performance of graph search-based algorithms highly depends on the complexity of the environment. This also causes the problem that a solution might not be found before the timeout is reached, which occurred here in \SI{3.4}{\percent} of all test cases. In summary, the CNN, whose output is qualitatively visualized in \cref{fig:cnn_predictions_all}, makes its predictions more reliably (no outages) with a lower latency. Both aspects are key features in safety-critical applications such as automated driving.

In order to better understand the effect of the different input grids on the performance of the CNN, an ablation study has been conducted. The results in \cref{tab:comparison_heuristics} show that removing features from the CNN's input causes a deterioration of at least one of the analyzed metrics. Furthermore, it can be seen that excluding the goal or the observations causes the greatest decline in performance. In contrast to that, the metrics only change slightly if the CNN is trained and evaluated without the obstacle or the unknown grid (not both). One of the reasons for this is that in many cases, both grids are complementary in the sense that the unknown grid allows to infer the obstacles and vice versa (see \cref{fig:cnn_inputs}).

The scenario-specific benchmark in \cref{tab:comparison_heuristics} highlights that the CNN's performance in the parking scenario is almost a factor two better compared to the maze-like structure arena. The resulting insight is that learning stationary sampling distributions in completely unstructured environments is a much harder task for the network than in semi-structured environments. This can also be seen in \cref{fig:cnn_predictions_all} on the \ifIEEE\else bottom \fi right, where the CNN prediction degenerates due to the complexity of the scenario. Possible reasons for this are longer prediction horizons, a larger variety of feasible maneuvers, and potentially heavier occlusions. It is left for future work to determine which network structure is the most suitable one for such challenging environments.

A visualization of the CNN prediction in two novel scenarios, namely a construction zone and and a cluttered 4-way intersection, can be found in the \ifIEEE\else lower left images \fi of \cref{fig:cnn_predictions_all}. Both scenarios were not seen during training and contain novel artifacts such as construction barrels and rectangular dumpsters. While the construction zone only requires the vehicle to follow the lane, the maneuver at the cluttered 4-way intersection consists of three steps: (1) exit the tight parking spot, (2) avoid the dumpster at the side of the road, and (3) make a turn at the intersection. The illustrated predictions showcase the models capability to deal with so far unseen artifacts as well as to generalize to novel scenarios. Only in the second case, the initial path prediction overlaps with the barriers in front of the vehicle. However, this is not a safety issue if the predictions are combined with a motion planner as highlighted in the next section.

\section{Guided Motion Planning} \label{sec:guided_mp}

This section evaluates the performance gain of the sampling-based motion planner BiRRT* guided by the pose predictions of the CNN. As before, the results are compared with purely uniform sampling and the OSE approach. The benchmark is conducted using Gazebo and ROS in three challenging automated driving scenarios that were excluded from the dataset in the previous section. As it can be seen in \cref{fig:scenarios_test}, these scenarios differ from the ones used to train the CNN (see \cref{fig:scenarios_train}) with respect to the spatial arrangement of the objects. Additionally, they contain novel artifacts like traffic cones and previously unseen vehicle geometries.

\ifIEEE
\else 
The parameters of the ego-vehicle are based on a full-size car and can be found in \cref{tab:vehicle_param}. Both the maximum curvature and the maximum curvature rate already include \SI{10}{\percent} control reserve for closed-loop execution.
\begin{table}[htbp]
	\caption{Parameters of the ego-vehicle.}
	\label{tab:vehicle_param}
	\centering
	\begin{tabular}{lll}
		\toprule
		parameter & value \\
		\midrule
		length  & \SI{4.926}{\metre} \\ 
		width & \SI{2.086}{\metre} \\
		wheel base & \SI{2.912}{\metre} \\
		maximum curvature & \SI{0.1982}{\per\metre} \\
		maximum curvature rate & \SI{0.1868}{\per\metre\squared} \\
		\bottomrule
	\end{tabular}
\end{table}
\fi

\begin{table*}[htbp]
	\vspace{0.25cm} %
	\caption{Motion planning results after \num{100}~runs of the same experiment in the three described scenarios. The table displays the comp. time for \num{100}~pose predictions, the time-to-first-solution TTFS, the number of vertices in the tree, the number of cusps, the path length, the cost of the first solution $\text{cost}_{\text{TTFS}}$, the cost after \SI{3}{\second} of optimization $\text{cost}_{\text{TTFS+3s}}$, and the success rate. Values given with mean and standard deviation are highlighted with a plus-minus sign.}
	\label{tab:exp_results}
	\centering
	\setlength{\tabcolsep}{4.9pt}  %
	\begin{tabular}{ccccccccccccc}
		\toprule
		scenario & heuristic & steer. fct. & pred.\,[\si{\milli\second}] & $\text{TTFS}\,[\si{\second}]$ & \#vertices\,$[\si{-}]$ & \#cusps\,$[\si{-}]$ & length\,$[\si{\meter}]$ & $\text{cost}_{\text{TTFS}}$\,$[\si{-}]$ & $\text{cost}_{\text{TTFS+3s}}$\,$[\si{-}]$ & success\,$[\si{\percent}]$ \\
		\midrule
		\multirow{3}{*}{I} & - & \multirow{3}{*}{{HC$^{00}$-RS}}
		& - & 0.57$\pm$0.41 & 79.2$\pm$14.3 & 4.0$\pm$1.2 & 35.2$\pm$4.8 & 162.0$\pm$48.3 & 124.1$\pm$28.2 & 100 \\ 
		& OSE & & 127.6 & 1.54$\pm$2.15 & 139.2$\pm$27.6 & 4.6$\pm$1.6 & 38.7$\pm$7.0 & 160.7$\pm$50.4 & 145.1$\pm$42.4 & 89 \\ 
		& CNN & & 82.8 & 0.13$\pm$0.07 & 140.6$\pm$7.5 & 2.2$\pm$0.8 & 28.1$\pm$3.9 & 134.4$\pm$37.2 & 93.4$\pm$18.9 & 100 \\
		\midrule
		\multirow{3}{*}{II} & - & \multirow{3}{*}{{CC$^{00}$-Dub.}}
		& - & 3.51$\pm$2.31 & 127.0$\pm$12.6 & 0.0$\pm$0.0 & 25.7$\pm$0.2 & 63.4$\pm$16.4 & 60.6$\pm$15.9 & 82 \\ 
		& OSE & & 17.4 & 0.12$\pm$0.14 & 187.2$\pm$12.0 & 0.0$\pm$0.0 & 25.7$\pm$0.1 & 50.1$\pm$11.2 & 37.4$\pm$2.9 & 100 \\ 
		& CNN & & 82.4 & 0.11$\pm$0.03 & 221.2$\pm$8.0 & 0.0$\pm$0.0 & 25.7$\pm$0.1 & 43.7$\pm$9.4 & 33.6$\pm$1.9 & 100 \\
		\midrule
		\multirow{3}{*}{III} & - & \multirow{3}{*}{{HC$^{00}$-RS}}			
		& - & 2.92$\pm$2.53 & 96.1$\pm$15.7 & 3.6$\pm$1.5 & 20.7$\pm$14.3 & 152.4$\pm$51.0 & 148.2$\pm$51.9 & 91 \\ 
		& OSE & & 19.0 & 0.02$\pm$0.03 & 154.6$\pm$6.0 & 3.0$\pm$1.0 & 12.8$\pm$1.5 & 143.6$\pm$21.0 & 119.2$\pm$16.3 & 100 \\ 
		& CNN & & 80.9 & 0.09$\pm$0.02 & 157.9$\pm$4.6 & 2.0$\pm$0.0 & 13.0$\pm$0.6 & 100.4$\pm$23.5 & 84.0$\pm$3.7 & 100 \\ 
		\bottomrule
	\end{tabular}
	\vspace{-0.25cm} %
\end{table*}

A \SI{60 x 60}{\meter} occupancy grid with a resolution of \SI{10}{\centi\meter} is used to represent the static environment. Collision checks are executed every \SI{10}{\centi\meter} using a convex polygon approximation of the vehicle's contour. The \num{20}~vertices of that polygon are inflated by \SI{10}{\centi\meter} serving as a hard safety buffer. An additional soft safety margin is introduced by converting the occupancy grid into a cost map with a \SI{25}{\centi\meter} inflation radius as illustrated in \cref{fig:scenarios_test}.

In order to compute an initial motion plan, BiRRT* is executed for maximal \SI{10}{\second}. Another \SI{3}{\second} are assigned for optimization if an initial solution is found. Otherwise, failure is reported for the corresponding run. Note that all computations are executed together with the simulation on an Intel Xeon E5@\SI{3.5}{\giga\hertz} and an Nvidia Quadro~K2200, which raises the CNN's computation time by a factor of two compared to the previous benchmark (see \cref{tab:comparison_heuristics} and \cref{tab:exp_results}).

Uniform sampling is conducted in the \SI{60 x 60}{\meter} region of the occupancy grid with a goal sampling frequency of \SI{5}{\percent}. For guided motion planning, heuristic samples are generated in batches of \num{100} at a frequency of \SI{4}{\hertz}. Each batch of samples is then evenly mixed with uniform samples such that the theoretical guarantees of BiRRT* are not violated.

The sampled vehicle poses are connected using the continuous curvature steering functions CC$^{00}$-Dubins~\cite{fraichard2004reeds} in Scenario~II and HC$^{00}$-Reeds-Shepp~\cite{banzhaf2017steer} in Scenario~I/III, where the superscript denotes the initial and final curvature of the steering procedure. The major difference between the two steering functions is that the first one constrains the vehicle to driving forwards only, while the second one allows moving forwards and backwards.

\begin{figure*}[htbp]
	\vspace{0.25cm} %
	\centering
	\newcommand{\subFigHeightConvergence}{4.55cm}
	\begin{subfigure}{0.312\textwidth}
		\centering	
		\includegraphics[height=\subFigHeightConvergence]{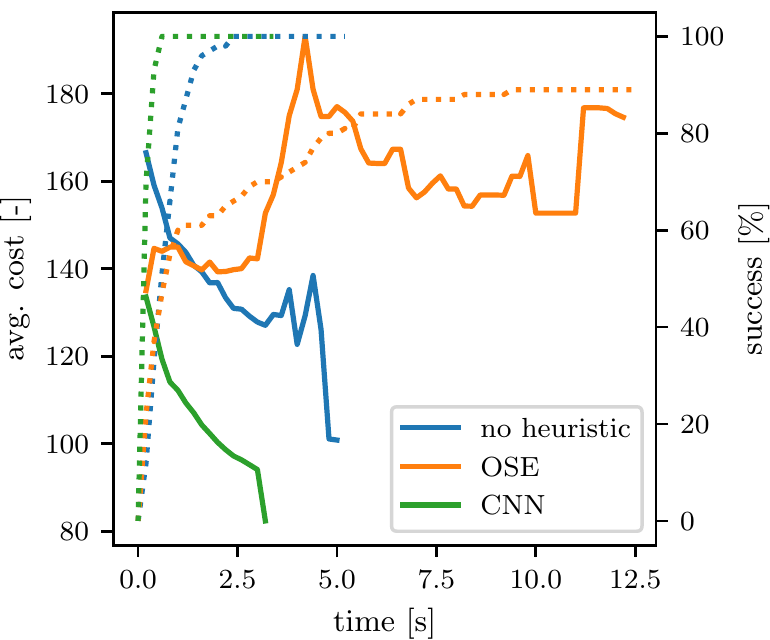}
		\subcaption{Scenario I}
	\end{subfigure}
	\begin{subfigure}{0.312\textwidth}
		\centering
		\includegraphics[height=\subFigHeightConvergence]{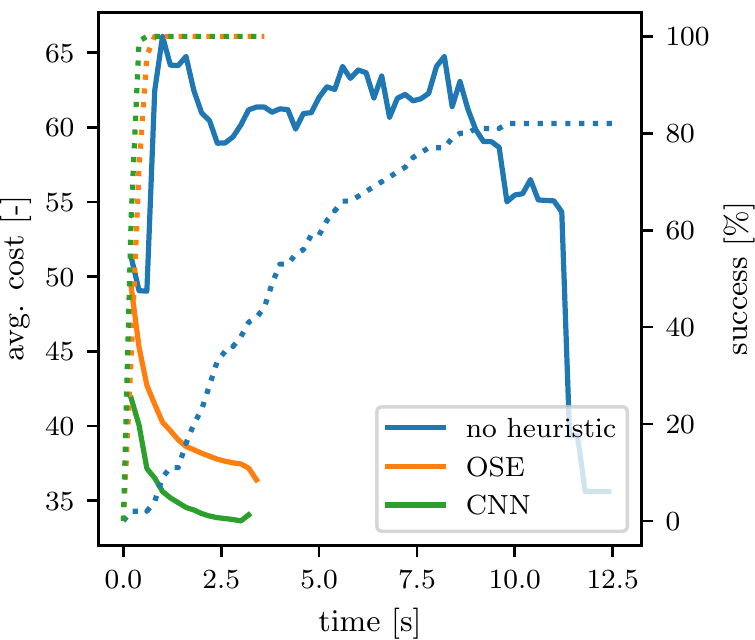}
		\subcaption{Scenario II}
	\end{subfigure}
	\begin{subfigure}{0.312\textwidth}
		\centering
		\includegraphics[height=\subFigHeightConvergence]{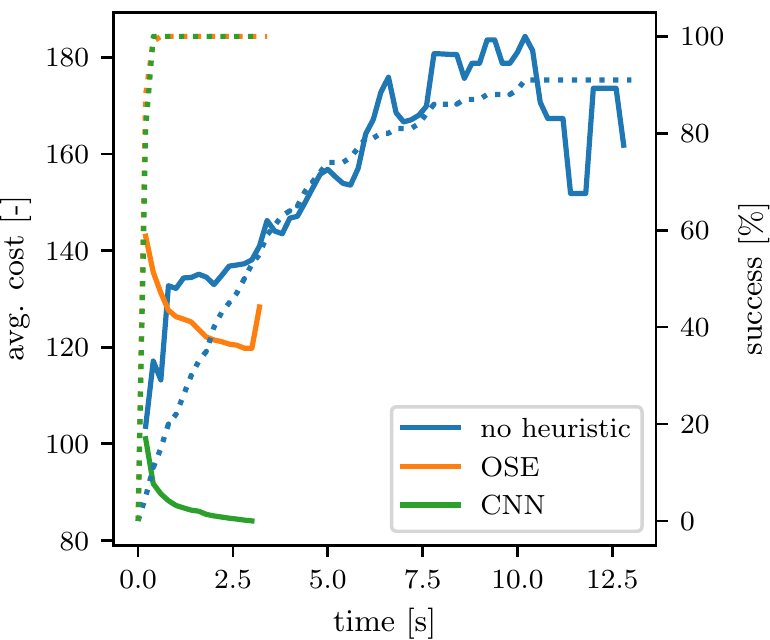}
		\subcaption{Scenario III}
	\end{subfigure}
	\caption{Comparison of the convergence rate in the three motion planning problems. The line colors distinguish the different heuristics, the solid lines illustrate the cost of the motion plan averaged over 100~runs, and the dotted lines display the percentage of succeeded runs. Note that while the cost of a single run is always monotonically decreasing, averaging over multiple runs might lead to a non-monotonic behavior.}
	\label{fig:convergance_comparison}
	\vspace{-0.25cm} %
\end{figure*}

A summary of the experimental results is shown in \cref{tab:exp_results}, where each setup is repeated \num{100}~times to compensate for randomization effects of the planner. It can be observed that BiRRT* in combination with the CNN-based predictions clearly outperforms the two other approaches. The CNN guides the motion planner to the best initial solution and helps to converge to the lowest cost. This is also highlighted in \cref{fig:convergance_comparison}, which illustrates the convergence rate of the different approaches in the tested scenarios. In addition to that, guiding BiRRT* with the CNN is the only approach with a success rate of \SI{100}{\percent} in all three scenarios. Compared to uniform sampling, the samples from the CNN not only stabilize the time-to-first solution (TTFS), but also reduce its mean by up to an order of magnitude. This value is currently limited by the network's inference time and can even be further reduced by a more advanced implementation of the prediction pipeline.

Guiding the motion planner with the OSE heuristic yields a low computation time in Scenario~II and III, where the circle expansion gives a good approximation of the planned path. As opposed to that, the performance deteriorates significantly in Scenario~I, where no solution can be found in \num{11} out of \num{100} runs. This is due to the fact that the OSE approach assumes a circular holonomic robot and only considers the nonholonomic constraints using cost terms~\cite{chen2015path}. As a result, the OSE proposes an immediate turnaround maneuver in Scenario~I, where an evasive multi-point turn would be required to resolve this situation (see \cref{fig:scenarios_test}).

\section{Conclusion} \label{sec:conc}

The complexity of motion planning combined with the real-time constraints of automated vehicles require heuristics that guarantee a fast convergence to a cost-minimizing solution. Within this context, a convolutional neural network~(CNN) is proposed that predicts ego-vehicle poses from a start to a goal state while taking into account the current surrounding. A benchmark on a recorded dataset highlights the CNN's capability to predict path corridors with a higher accuracy compared to two baselines: uniform sampling and a state-of-the-art A*-based approach. Furthermore, it is demonstrated that the CNN has the capability to adapt its prediction to novel scenarios with previously unseen artifacts. Together with the sampling-based motion planner BiRRT*, this results in a significant reduction of computation time to about \SI{100}{\milli\second}, high-quality paths, and success rates of \SI{100}{\percent} in three challenging automated driving scenarios. In conclusion, the proposed method is especially suitable for real-time motion planning in complex environments.

\section{Acknowledgement}

The authors thank Florian Faion for valuable discussions and insights.

\bibliographystyle{myIEEEtran}
\bibliographystyle{cite} 
\bibliography{root}

\begin{thebibliography}{10}
\providecommand{\url}[1]{#1}
\csname url@rmstyle\endcsname
\providecommand{\newblock}{\relax}
\providecommand{\bibinfo}[2]{#2}
\providecommand\BIBentrySTDinterwordspacing{\spaceskip=0pt\relax}
\providecommand\BIBentryALTinterwordstretchfactor{4}
\providecommand\BIBentryALTinterwordspacing{\spaceskip=\fontdimen2\font plus
\BIBentryALTinterwordstretchfactor\fontdimen3\font minus
  \fontdimen4\font\relax}
\providecommand\BIBforeignlanguage[2]{{%
\expandafter\ifx\csname l@#1\endcsname\relax
\typeout{** WARNING: IEEEtran.bst: No hyphenation pattern has been}%
\typeout{** loaded for the language `#1'. Using the pattern for}%
\typeout{** the default language instead.}%
\else
\language=\csname l@#1\endcsname
\fi
#2}}

\bibitem{banzhaf2019learning}
H.~Banzhaf \emph{et~al.}, ``{Learning to Predict Ego-Vehicle Poses for
  Sampling-Based Nonholonomic Motion Planning},'' \emph{IEEE Robotics and
  Automation Letters}, 2019, {DOI 10.1109/LRA.2019.2893975}.

\bibitem{dolgov2010path}
D.~Dolgov \emph{et~al.}, ``{Path Planning for Autonomous Vehicles in Unknown
  Semi-structured Environments},'' \emph{The International Journal of Robotics
  Research}, 2010.

\bibitem{chen2015path}
C.~Chen \emph{et~al.}, ``{Path Planning with Orientation-Aware Space
  Exploration Guided Heuristic Search for Autonomous Parking and
  Maneuvering},'' in \emph{IEEE Intelligent Vehicles Symposium}, 2015.

\bibitem{gonzalez2016review}
D.~Gonz{\'a}lez \emph{et~al.}, ``{A Review of Motion Planning Techniques for
  Automated Vehicles},'' \emph{IEEE Transactions on Intelligent Transportation
  Systems}, 2016.

\bibitem{jordan2013optimal}
M.~Jordan and A.~Perez, ``{Optimal Bidirectional Rapidly-Exploring Random
  Trees},'' MIT, Tech. Rep., 2013, {TR 2013-021}.

\bibitem{likhachev2009planning}
M.~Likhachev and D.~Ferguson, ``{Planning Long Dynamically-Feasible Maneuvers
  for Autonomous Vehicles},'' \emph{The International Journal of Robotics
  Research}, 2009.

\bibitem{pomerleau1989alvinn}
D.~A. Pomerleau, ``{ALVINN: An Autonomous Land Vehicle In a Neural Network},''
  in \emph{Advances in Neural Information Processing Systems}, 1989.

\bibitem{bojarski2016end}
M.~Bojarski \emph{et~al.}, ``{End to End Learning for Self-Driving Cars},''
  \emph{arXiv preprint arXiv:1604.07316}, 2016.

\bibitem{tamar2017learning}
A.~Tamar \emph{et~al.}, ``{Learning from the Hindsight Plan - Episodic MPC
  Improvement},'' in \emph{IEEE International Conference on Robotics and
  Automation}, 2017.

\bibitem{kim2018guiding}
B.~Kim, L.~P. Kaelbling, and T.~Lozano-P{\'e}rez, ``{Guiding Search in
  Continuous State-action Spaces by Learning an Action Sampler from Off-target
  Search Experience},'' in \emph{AAAI Conference on Artificial Intelligence},
  2018.

\bibitem{goodfellow2014generative}
I.~Goodfellow \emph{et~al.}, ``{Generative Adversarial Nets},'' in
  \emph{Advances in Neural Information Processing Systems}, 2014.

\bibitem{choudhury2018data}
S.~Choudhury \emph{et~al.}, ``{Data-driven planning via imitation learning},''
  \emph{The International Journal of Robotics Research}, 2018.

\bibitem{hubschneider2017integrating}
C.~Hubschneider \emph{et~al.}, ``{Integrating End-to-End Learned Steering into
  Probabilistic Autonomous Driving},'' in \emph{IEEE International Conference
  on Intelligent Transportation Systems}, 2017.

\bibitem{sohn2015learning}
K.~Sohn, H.~Lee, and X.~Yan, ``{Learning Structured Output Representation using
  Deep Conditional Generative Models},'' in \emph{Advances in Neural
  Information Processing Systems}, 2015.

\bibitem{ichter2018learning}
B.~Ichter, J.~Harrison, and M.~Pavone, ``{Learning Sampling Distributions for
  Robot Motion Planning},'' in \emph{IEEE International Conference on Robotics
  and Automation}, 2018.

\bibitem{perez2018learning}
N.~P{\'e}rez-Higueras \emph{et~al.}, ``{Learning Human-Aware Path Planning with
  Fully Convolutional Networks},'' in \emph{IEEE International Conference on
  Robotics and Automation}, 2018.

\bibitem{karaman2010incremental}
S.~Karaman and E.~Frazzoli, ``{Incremental Sampling-based Algorithms for
  Optimal Motion Planning},'' \emph{Robotics Science and Systems VI}, 2010.

\bibitem{qureshi2018motion}
A.~H. Qureshi \emph{et~al.}, ``{Motion Planning Networks},'' \emph{arXiv
  preprint arXiv:1806.05767}, 2018.

\bibitem{ichter2018robot}
B.~Ichter and M.~Pavone, ``{Robot Motion Planning in Learned Latent Spaces},''
  \emph{arXiv preprint arXiv:1807.10366}, 2018.

\bibitem{srinivas2018universal}
A.~Srinivas \emph{et~al.}, ``{Universal Planning Networks},'' in
  \emph{International Conference on Machine Learning}, 2018.

\bibitem{baumann2018path}
U.~Baumann \emph{et~al.}, ``{Predicting Ego-Vehicle Paths from Environmental
  Observations with a Deep Neural Network},'' in \emph{IEEE International
  Conference on Robotics and Automation}, 2018.

\bibitem{caltagirone2017lidar}
L.~Caltagirone \emph{et~al.}, ``{LIDAR-based Driving Path Generation Using
  Fully Convolutional Neural Networks},'' in \emph{IEEE International
  Conference on Intelligent Transportation Systems}, 2017.

\bibitem{badrinarayanan2015segnet}
V.~Badrinarayanan \emph{et~al.}, ``{SegNet: A Deep Convolutional
  Encoder-Decoder Architecture for Image Segmentation},'' \emph{arXiv preprint
  arXiv:1511.00561}, 2015.

\bibitem{ioffe2015batch}
S.~Ioffe and C.~Szegedy, ``{Batch Normalization: Accelerating Deep Network
  Training by Reducing Internal Covariate Shift},'' in \emph{International
  Conference on Machine Learning}, 2015.

\bibitem{nair2010rectified}
V.~Nair and G.~E. Hinton, ``{Rectified Linear Units Improve Restricted
  Boltzmann Machines},'' in \emph{International Conference on Machine
  Learning}, 2010.

\bibitem{thrun2005probabilistic}
S.~Thrun \emph{et~al.}, \emph{{Probabilistic Robotics}}.\hskip 1em plus 0.5em
  minus 0.4em\relax MIT Press, 2005.

\bibitem{kingma2015adam}
D.~P. Kingma and J.~L. Ba, ``{Adam: A Method for Stochastic Optimization},'' in
  \emph{International Conference on Learning Representations}, 2015.

\bibitem{banzhaf2017deques}
H.~Banzhaf \emph{et~al.}, ``{High Density Valet Parking Using $k$-Deques in
  Driveways},'' in \emph{IEEE Intelligent Vehicles Symposium}, 2017.

\bibitem{fraichard2004reeds}
T.~Fraichard and A.~Scheuer, ``{From Reeds and Shepp's to Continuous-Curvature
  Paths},'' \emph{IEEE Transactions on Robotics}, 2004.

\bibitem{banzhaf2017steer}
H.~Banzhaf \emph{et~al.}, ``{Hybrid Curvature Steer: A Novel Extend Function
  for Sampling-Based Nonholonomic Motion Planning in Tight Environments},'' in
  \emph{IEEE International Conference on Intelligent Transportation Systems},
  2017.

\end{thebibliography}

\end{document}